\documentclass[sigconf]{acmart}

\usepackage{amsmath}

\usepackage{lineno}
\usepackage{setspace}
\usepackage{graphicx} 
\usepackage{hyperref} 
\usepackage[capitalize]{cleveref}
\usepackage{textcomp}
\usepackage{balance}
\usepackage{booktabs}
\usepackage{subcaption}
\usepackage{setspace} 
\usepackage{geometry} 
\usepackage{fancyhdr} 
\usepackage{multirow}
\usepackage{svg}

\usepackage{pifont}

\AtBeginDocument{%
  }

\copyrightyear{2026}
\acmYear{2026}
\setcopyright{cc}
\setcctype{by}
\acmConference[MAD '26]{The 5th ACM International Workshop on Multimedia AI against Disinformation }{June 16--19, 2026}{Amsterdam, Netherlands}
\acmBooktitle{The 5th ACM International Workshop on Multimedia AI against Disinformation (MAD '26), June 16--19, 2026, Amsterdam, Netherlands}
\acmDOI{10.1145/3810988.3812665}
\acmISBN{979-8-4007-2700-9/2026/06}

\acmSubmissionID{4}

\begin{document}

\title{Navigating the Challenges of AI-Generated Image Detection in the Wild: What Truly Matters?}

\author{Despina Konstantinidou}
\email{dekonstantinidou@gmail.com}
\affiliation{%
  \institution{Information Technologies Institute - Centre for Research and Technology Hellas}
  \city{Thessaloniki}
  \country{Greece}
}
\author{Dimitrios Karageorgiou}
\email{dkarageo@iti.gr}
\affiliation{%
  \institution{Information Technologies Institute - Centre for Research and Technology Hellas}
  \city{Thessaloniki}
  \country{Greece}
}
\author{Christos Koutlis}
\email{ckoutlis@iti.gr}
\affiliation{%
  \institution{Information Technologies Institute - Centre for Research and Technology Hellas}
  \city{Thessaloniki}
  \country{Greece}
}
\author{Olga Papadopoulou}
\email{olgapapa@iti.gr}
\affiliation{%
  \institution{Information Technologies Institute - Centre for Research and Technology Hellas}
  \city{Thessaloniki}
  \country{Greece}
}
\author{Emmanouil Schinas}
\email{manosetro@iti.gr}
\affiliation{%
  \institution{Information Technologies Institute - Centre for Research and Technology Hellas}
  \city{Thessaloniki}
  \country{Greece}
}
\author{Symeon Papadopoulos}
\email{papadop@iti.gr}
\affiliation{%
  \institution{Information Technologies Institute - Centre for Research and Technology Hellas}
  \city{Thessaloniki}
  \country{Greece}
}








\renewcommand{\shortauthors}{Konstantinidou et al.}

\begin{abstract}
As generative Artificial Intelligence (AI) advances, the realism of AI generated imagery has reached a threshold capable of deceiving even vigilant human observers. Yet, while current AI-generated Image Detection (AID) approaches perform exceptionally well on controlled benchmark datasets, they struggle significantly with real-world cases. To study this behavior we introduce the \textbf{ITW-SM} dataset, a curated collection of real and AI-generated images originating from major social media platforms. We employ it to analyze the effects of key design choices typically considered when building a detector, involving its architecture, pre-trained latent spaces, training data as well as pre-processing approaches. We indicate that naively scaling the pre-training stage or opting for more training data does not always lead to better detection performance. Instead, our work reveals that it is crucial to optimize each design choice to enable the processing pipeline to propagate and effectively analyze both low-level traces as well as high-level image semantics. Building on our findings, we achieve a substantial average improvement of 26.87\% in AUC across multiple state-of-the-art detection approaches and under real-world conditions, providing a roadmap for developing more resilient detectors. Our assets are available on \href{https://mever-team.github.io/itw-sm}{\textcolor{blue}{https://mever-team.github.io/itw-sm}}.
\end{abstract}




\maketitle

\section{Introduction}
Generative AI has revolutionized digital media, enabling the creation of photorealistic content on demand via natural language descriptions~\cite{li2025comprehensive}. While such tools offer immense creative potential, they also pose significant risks, as they can be maliciously exploited to spread disinformation, facilitate impersonation, or enable fraudulent activities~\cite{tredinnick2023dangers}. Because their high fidelity deceives even careful human observers~\cite{lu2023seeingbelievingbenchmarkinghuman, papa2023ontheuseofstablediffusion}, and the sheer volume of online media prevents manual review, robust automated AI-generated Image Detection (AID) deployed in the wild is crucial.

Existing AID approaches span pixel-level methods operating on raw image data~\cite{wang2020cnngeneratedimagessurprisinglyeasy,gragnaniello2021gangeneratedimageseasy,corvi2022detectionsyntheticimagesgenerated,Dogoulis_2023}, fingerprint-based methods targeting frequency domain or reconstruction artifacts~\cite{bammey2023synthbuster,durall2020watchupconvolutioncnnbased,yanhao2024masksim, karageorgiou2024anyresolutionaigeneratedimagedetection}, and zero-shot approaches for generalized, training-free detection~\cite{he2024rigidtrainingfreemodelagnosticframework,cozzolino2024zeroshotdetectionaigeneratedimages}. Despite the growing number of AID methods, achieving robust performance in real-world scenarios remains a challenge, as most AID models perform exceptionally well on benchmark datasets~\cite{schinas2024sidbenchpythonframeworkreliably}, which are typically generated in controlled environments, but collapse when unconditionally tested on content shared online~\cite{karageorgiou2024evolution, corvi2023intriguingpropertiessyntheticimages,cozzolino2024raisingbaraigeneratedimage}. This disparity in performance highlights the need for a systematic study of the factors influencing AID robustness in real-world settings. Through our analysis, we identify four factors that significantly affect the performance of detectors in the wild:

\begin{enumerate}
    \item \textbf{Training data}: 
    In AID, the challenge lies in ensuring that the training data reflects the diversity and complexity of real-world cases. If the training distribution \( \mathcal{P}_{\text{train}}(x) \) is derived from controlled and limited generators, it may diverge from the actual distribution encountered at deployment \( \mathcal{P}_{\text{actual}}(x) \). 
        
    \item \textbf{Pre-trained latent space}: A backbone maps input images \( x \) to a latent space through a function $\mathcal{B}(x)$, to extract discriminative features that expose generative artifacts~\cite{bammey2023synthbuster}. Its effectiveness can be quantified by the expected classification loss $
    L_{\mathcal{B}} = \mathbb{E}_{x \sim \mathcal{P}_{\text{actual}}} \left[ \ell\left( G(\mathcal{B}(x)), y \right) \right]$ 
    where \( \ell \) is a classification loss function, $G$ is a projection function mapping features to the decision space and \( y \) is the true label. Lower \( L_{\mathcal{B}} \) implies a more effective backbone for modeling $\mathcal{P}_{\text{actual}}$. \looseness=-1  
    
    \item \textbf{Pre-processing methods}: These play a crucial role in ensuring that models can handle input data efficiently, as popular computer vision models, like convolutional neural networks~\cite{krizhevsky2012imagenet} and vision transformers~\cite{dosovitskiy2020image}, scale quadratically with image size. Instead of the typical 224 $\times$ 224 image size, images in the wild can be several megapixels large. With resizing being considered detrimental for AID—due to its tendency to erase subtle high-frequency traces left by the generation process—cropping techniques become essential~\cite{konstantinidou2025texturecropenhancingsyntheticimage, karageorgiou2024anyresolutionaigeneratedimagedetection}. By cropping images into smaller parts through a function $\mathcal{C}(x)$, models can analyze the parts individually and focus on localized details that may carry important information about the presence of synthesis artifacts.  

    \item \textbf{Data augmentations}: These are crucial for improving model robustness, as they simulate real-world variations. Data augmentations can be seen as a set of transformations \( \mathcal{T} \) applied to the training images, where each transformation \( t \in \mathcal{T} \) maps an image $x \sim \mathcal{P}_{\text{train}}(x)$ to a new image \( t(x) \). By introducing perturbations similar to those encountered online, the model is exposed to a wider range of potential inputs, improving its generalization capabilities. 
\end{enumerate}

Our study systematically evaluates these components to identify optimal strategies for improving AID robustness in the wild. In particular, each of these factors contribute to the overall expected error in AID, which can be expressed as \( \epsilon_{\text{AID}} = g(\mathcal{P}_{\text{train}}, \mathcal{B}, \mathcal{C}, \mathcal{T}) \), where $g$ is a function that models their interactions. Our work aims to provide insights regarding such interactions. To facilitate a realistic evaluation setup and address gaps in existing resources, we introduce \textbf{ITW-SM}, a new in-the-wild test dataset of 10,000 real and generated images collected from popular social media platforms. Using ITW-SM in tandem with established benchmarks, we provide actionable insights into why current methods fail in the wild and how to improve them. Our contributions include:

\begin{itemize}
    \item A systematic experimental evaluation revealing the strengths and weaknesses of various AID approaches in the wild.
    \item The introduction of \textbf{ITW-SM}, a new in-the-wild AID benchmark dataset collected from four popular social media platforms, designed to support evaluation under realistic and unconstrained conditions.
    \item An impact analysis of training data, pre-trained latent spaces, model architectures, pre-processing stages and data augmentations on AID performance in the wild. 
    \item An average improvement of $26.87\%$ in AUC across four types of detectors and under real-world conditions.
    \item A set of recommendations for designing more robust AID models capable of handling in-the-wild variations.
\end{itemize}
\section{Related Work}
\label{sec:related}

\begin{figure*}
    \begin{subfigure}[b]{0.19\textwidth}
        \includegraphics[width=\textwidth]{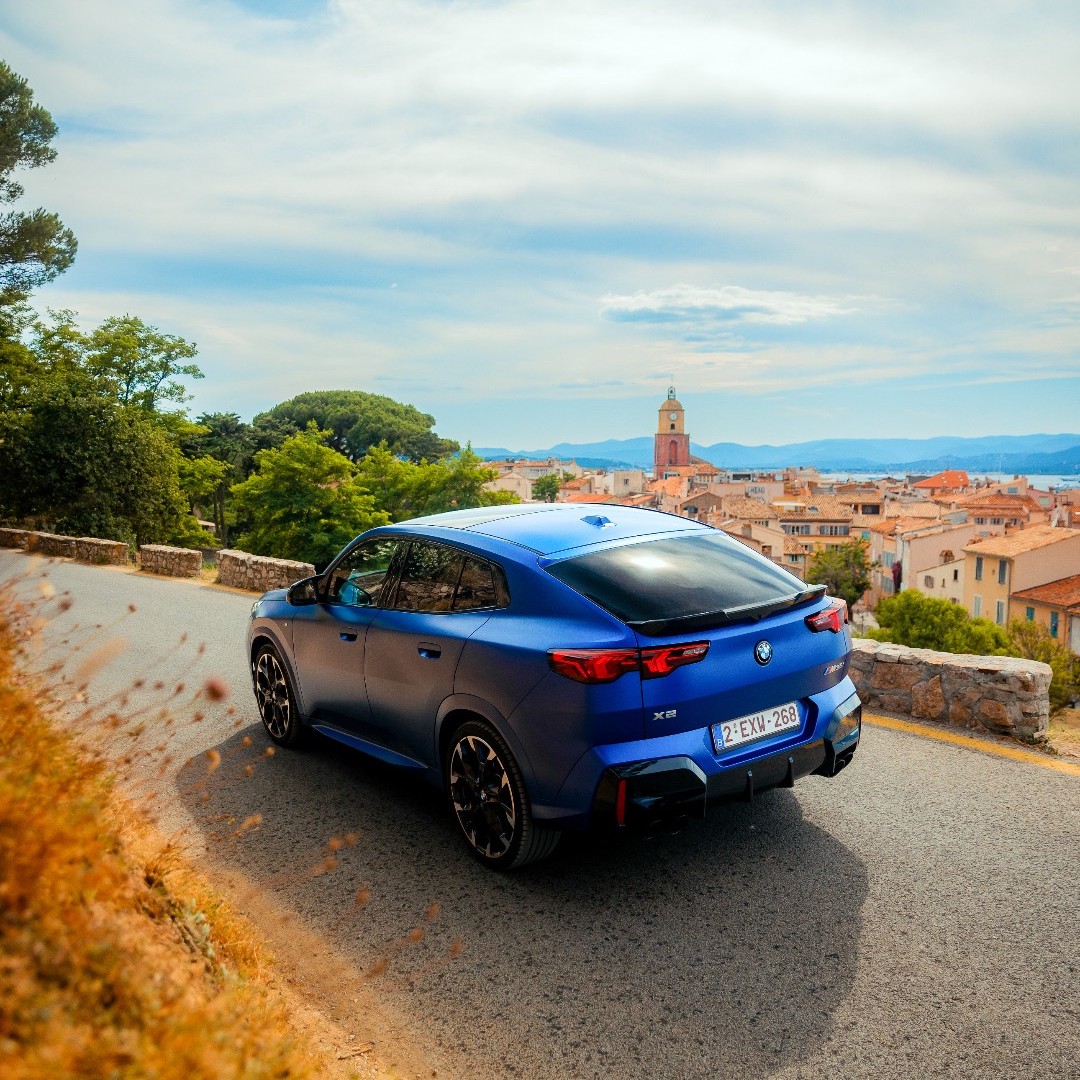}
        \caption{}
    \end{subfigure}
    \hfill
    \begin{subfigure}[b]{0.19\textwidth}
        \includegraphics[width=\textwidth]{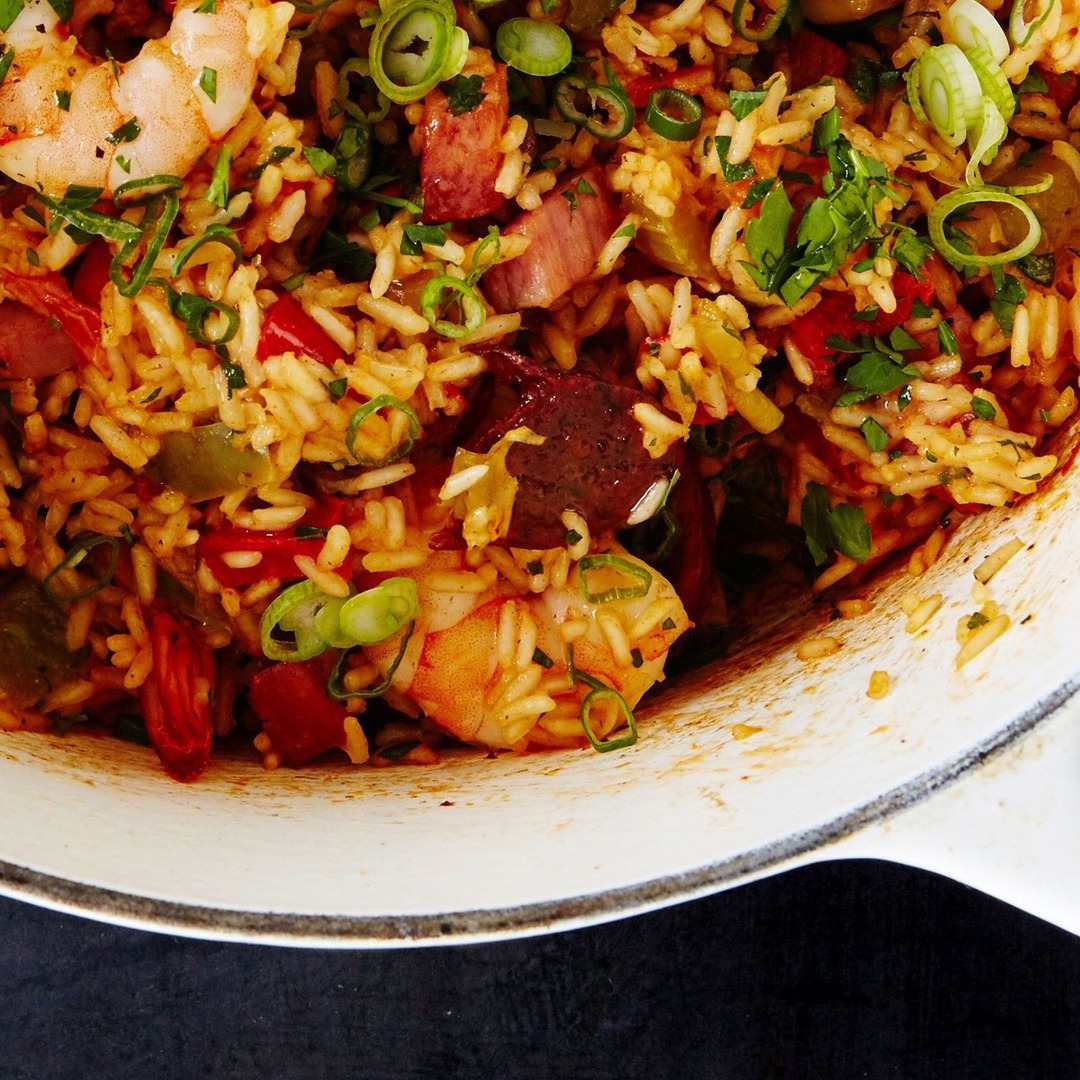}
        \caption{}
    \end{subfigure}
    \hfill
    \begin{subfigure}[b]{0.19\textwidth}
        \includegraphics[width=\textwidth]{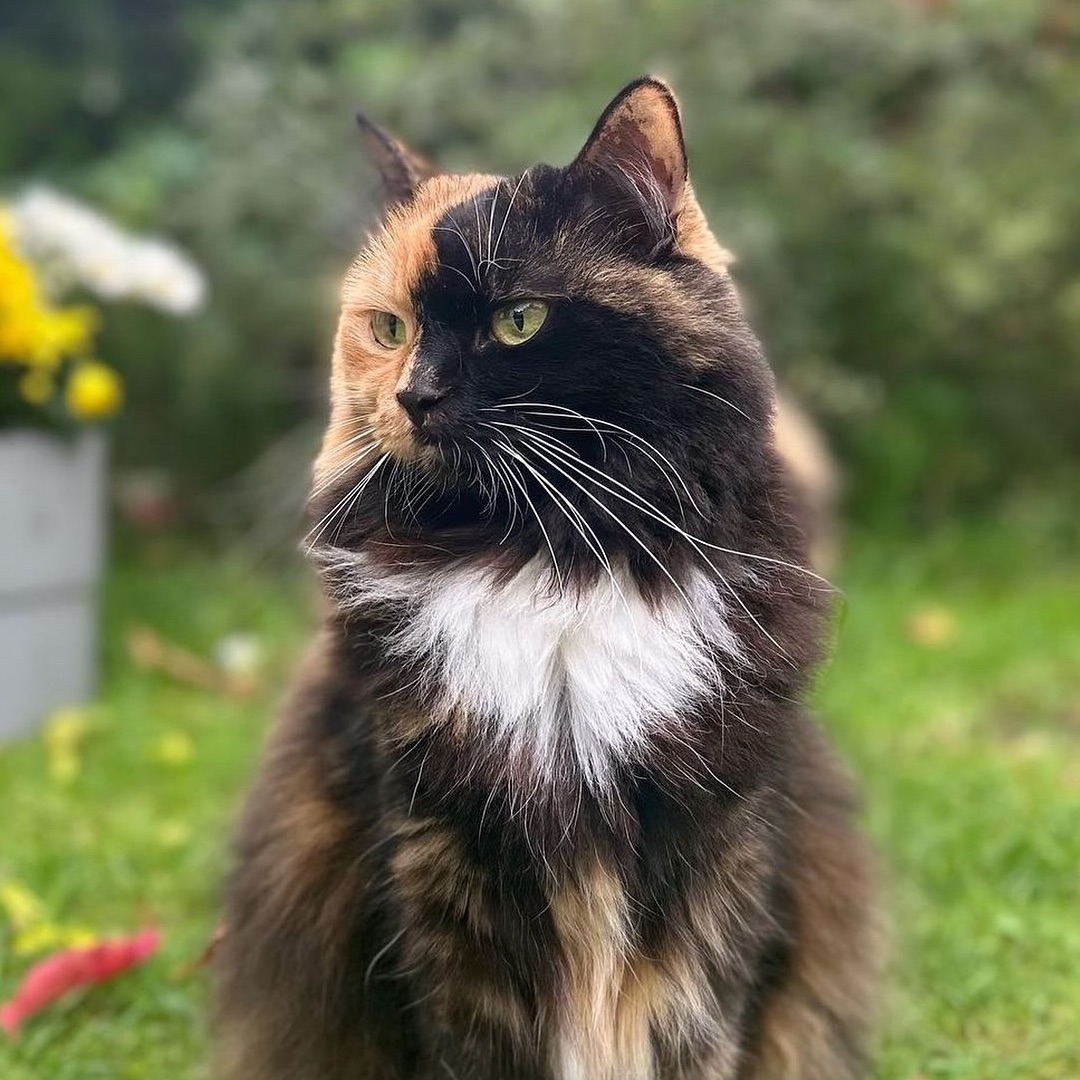}
        \caption{}
    \end{subfigure}
    \hfill
    \begin{subfigure}[b]{0.19\textwidth}
        \includegraphics[width=\textwidth]{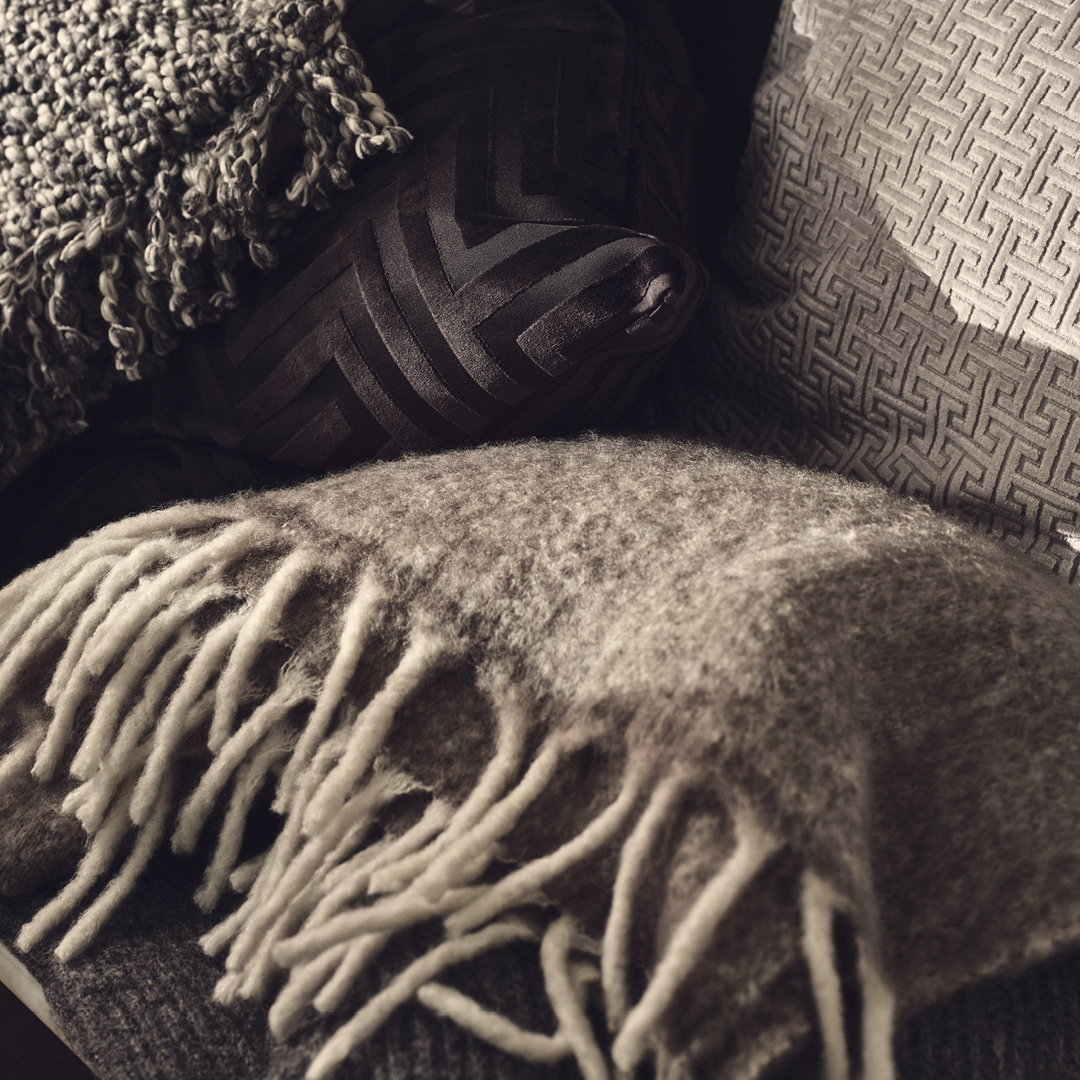}
        \caption{}
    \end{subfigure}
    \hfill
    \begin{subfigure}[b]{0.19\textwidth}
        \includegraphics[width=\textwidth]{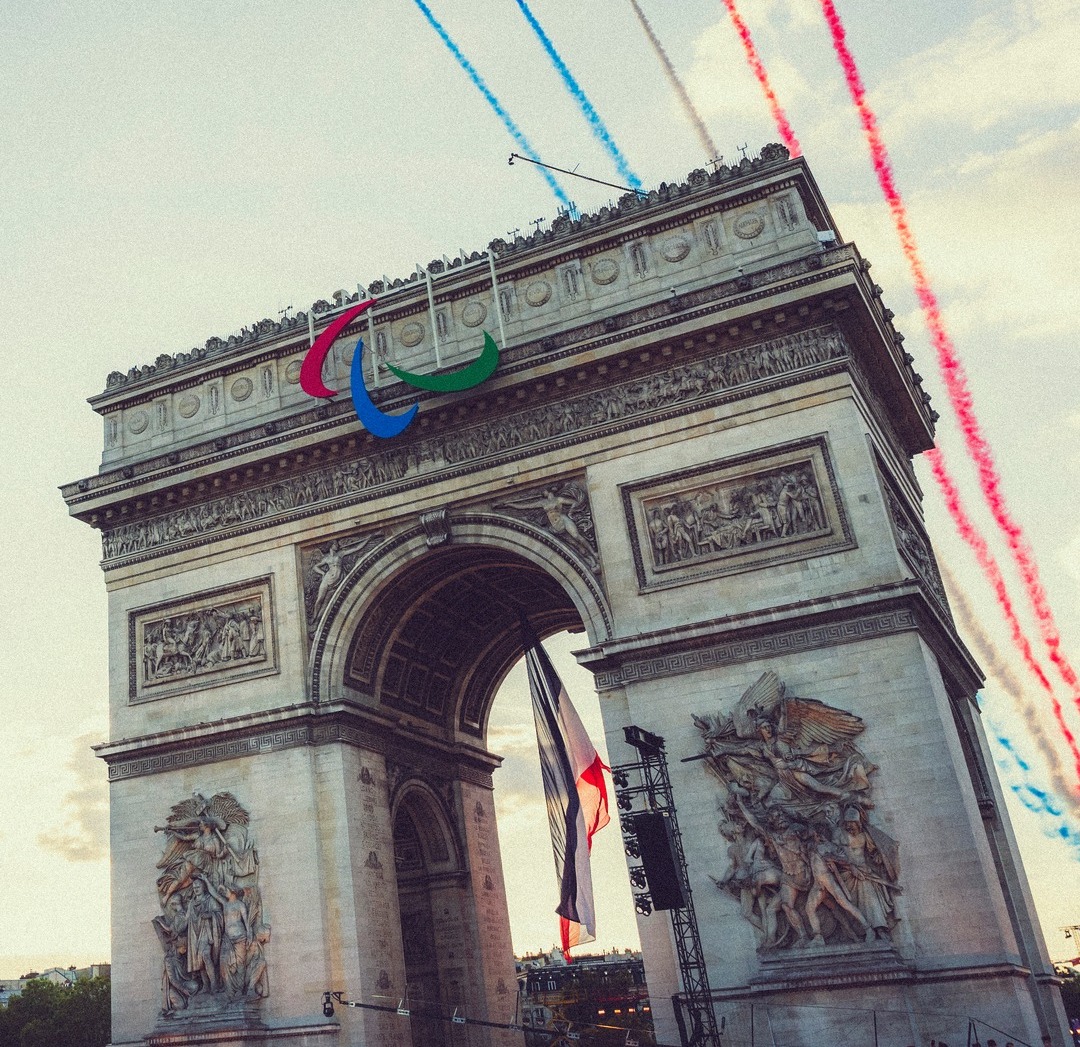}
        \caption{}
    \end{subfigure}
    \hfill
    \begin{subfigure}[b]{0.19\textwidth}
        \includegraphics[width=\textwidth]{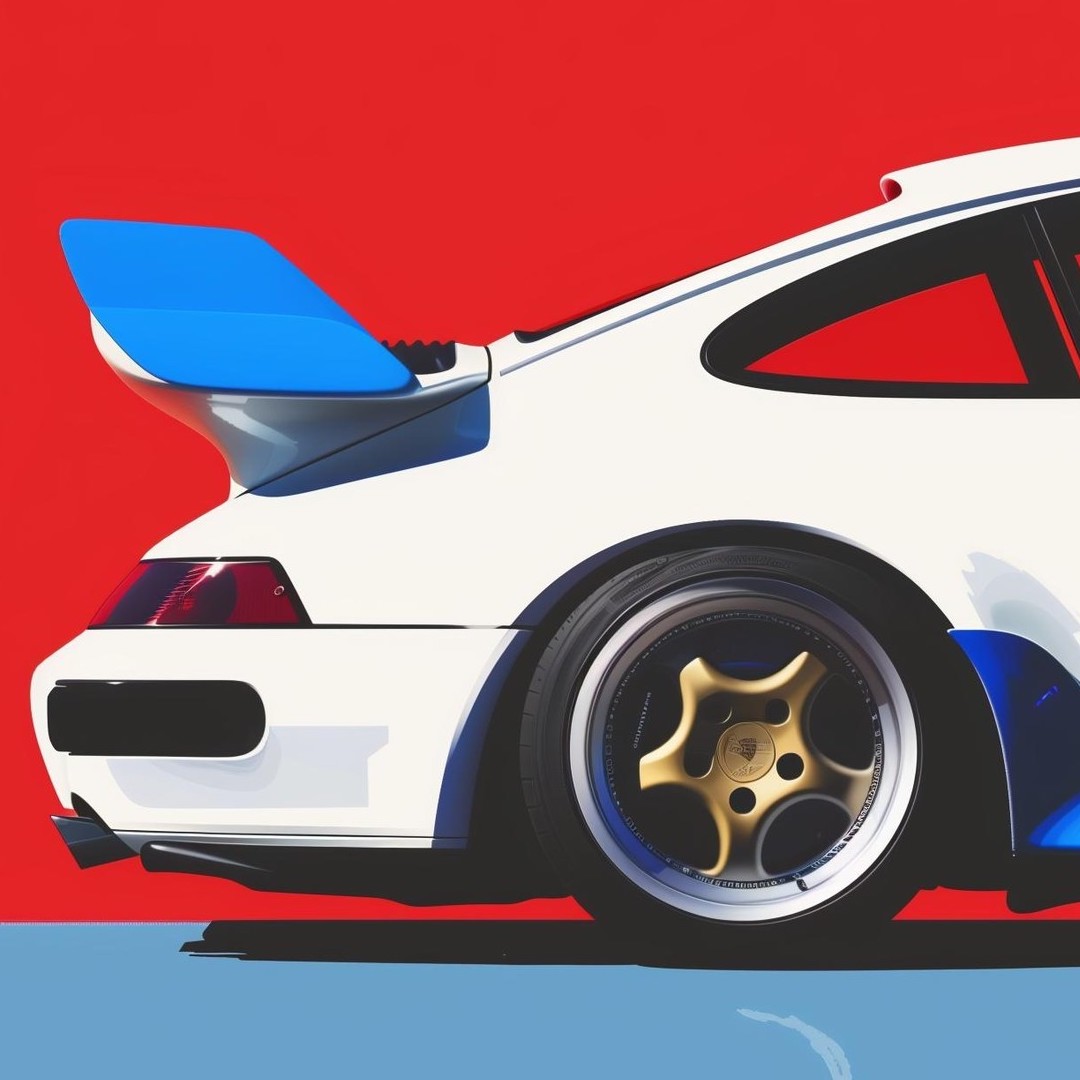}
        \caption{}
    \end{subfigure}
    \hfill
    \begin{subfigure}[b]{0.19\textwidth}
        \includegraphics[width=\textwidth]{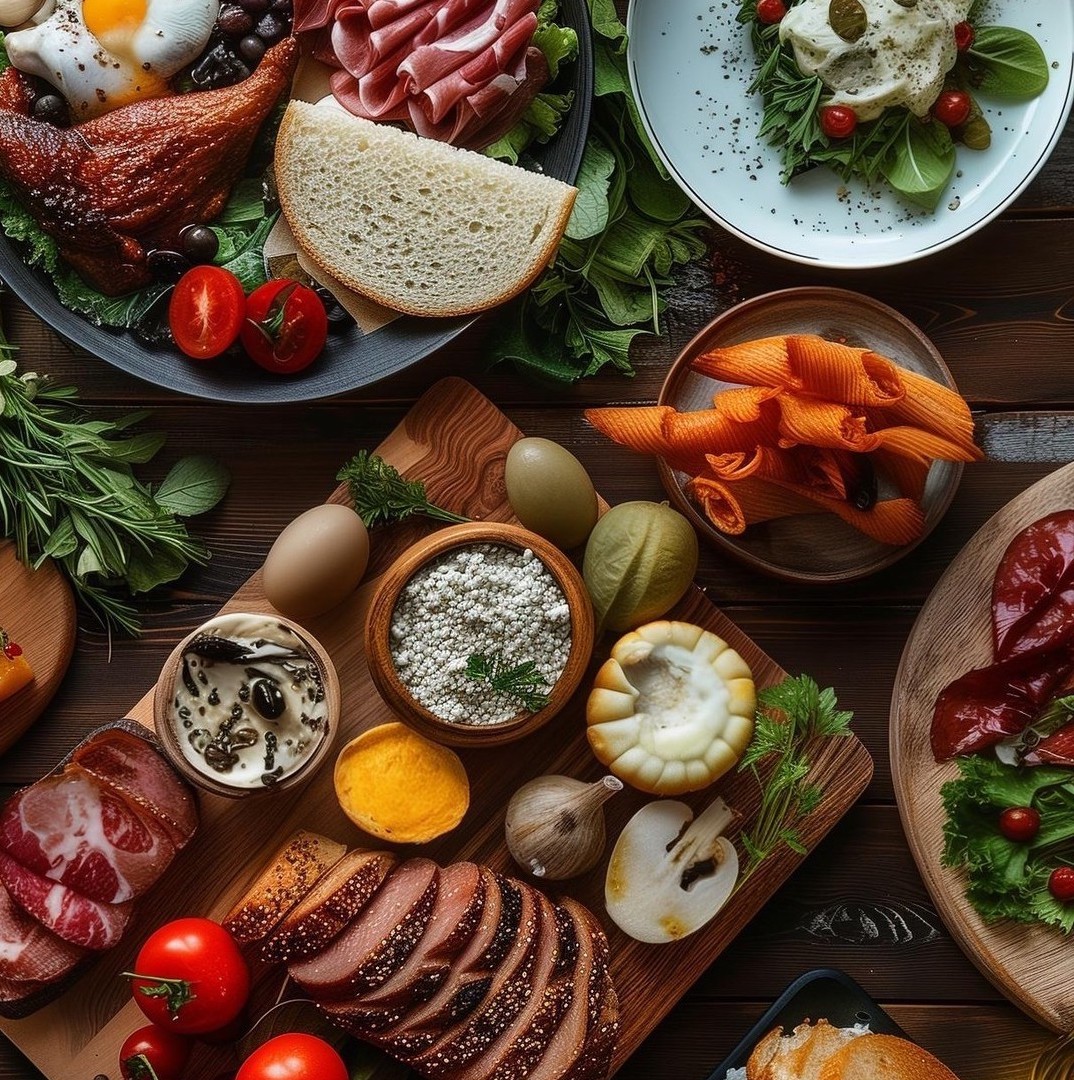}
        \caption{}
    \end{subfigure}
    \hfill
    \begin{subfigure}[b]{0.19\textwidth}
        \includegraphics[width=\textwidth]{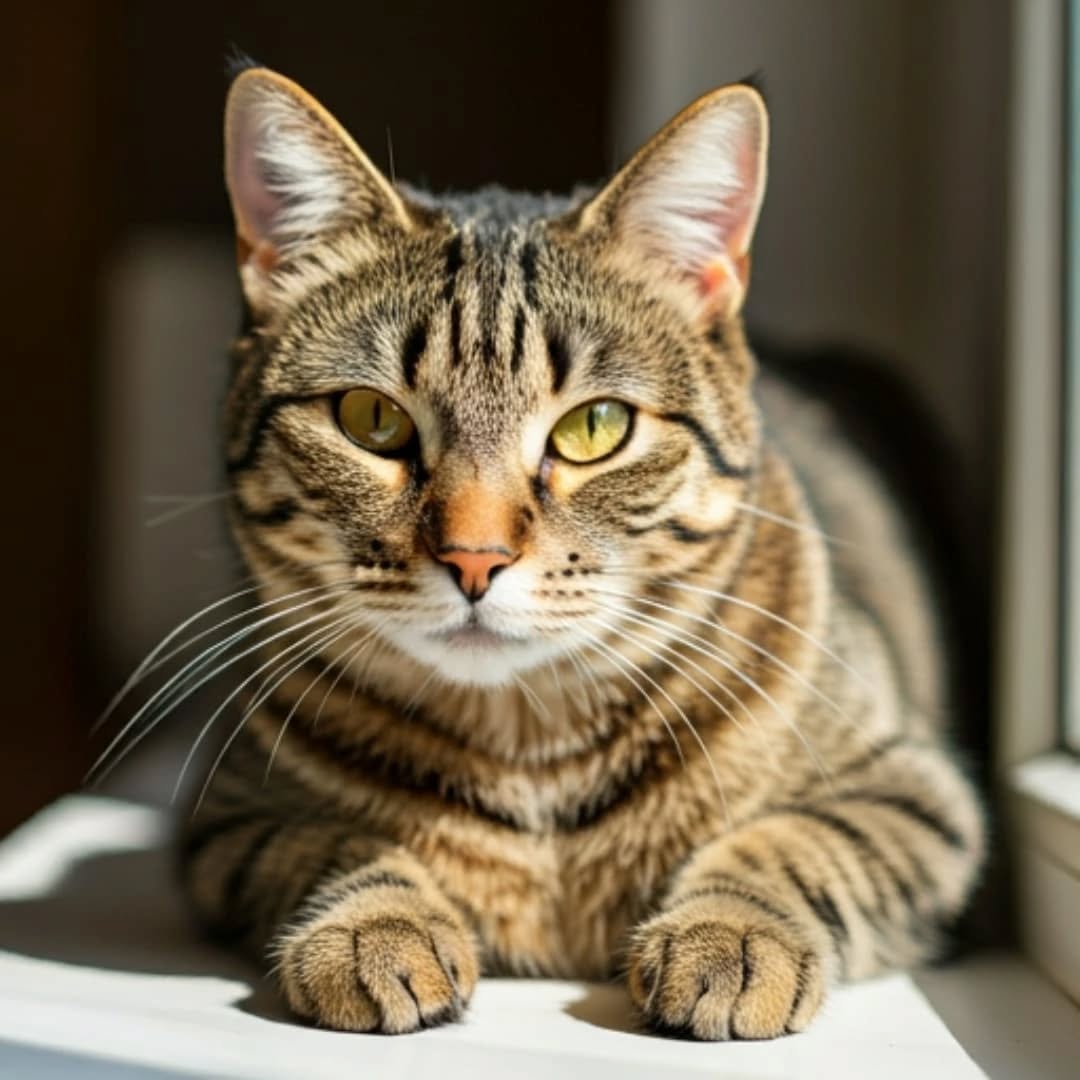}
        \caption{}
    \end{subfigure}
    \begin{subfigure}[b]{0.19\textwidth}
        \includegraphics[width=\textwidth]{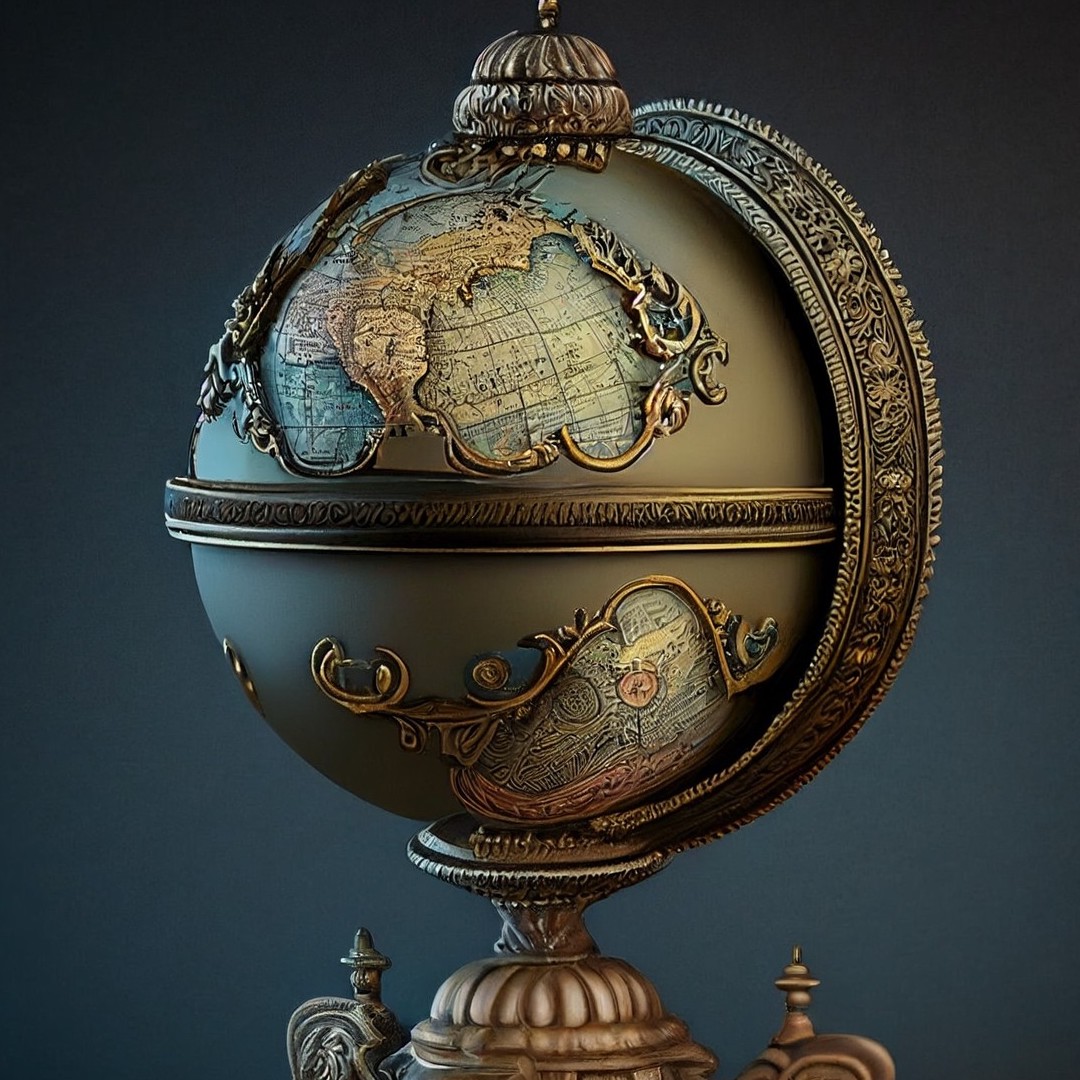}
        \caption{}
    \end{subfigure}
    \hfill
    \begin{subfigure}[b]{0.19\textwidth}
        \includegraphics[width=\textwidth]{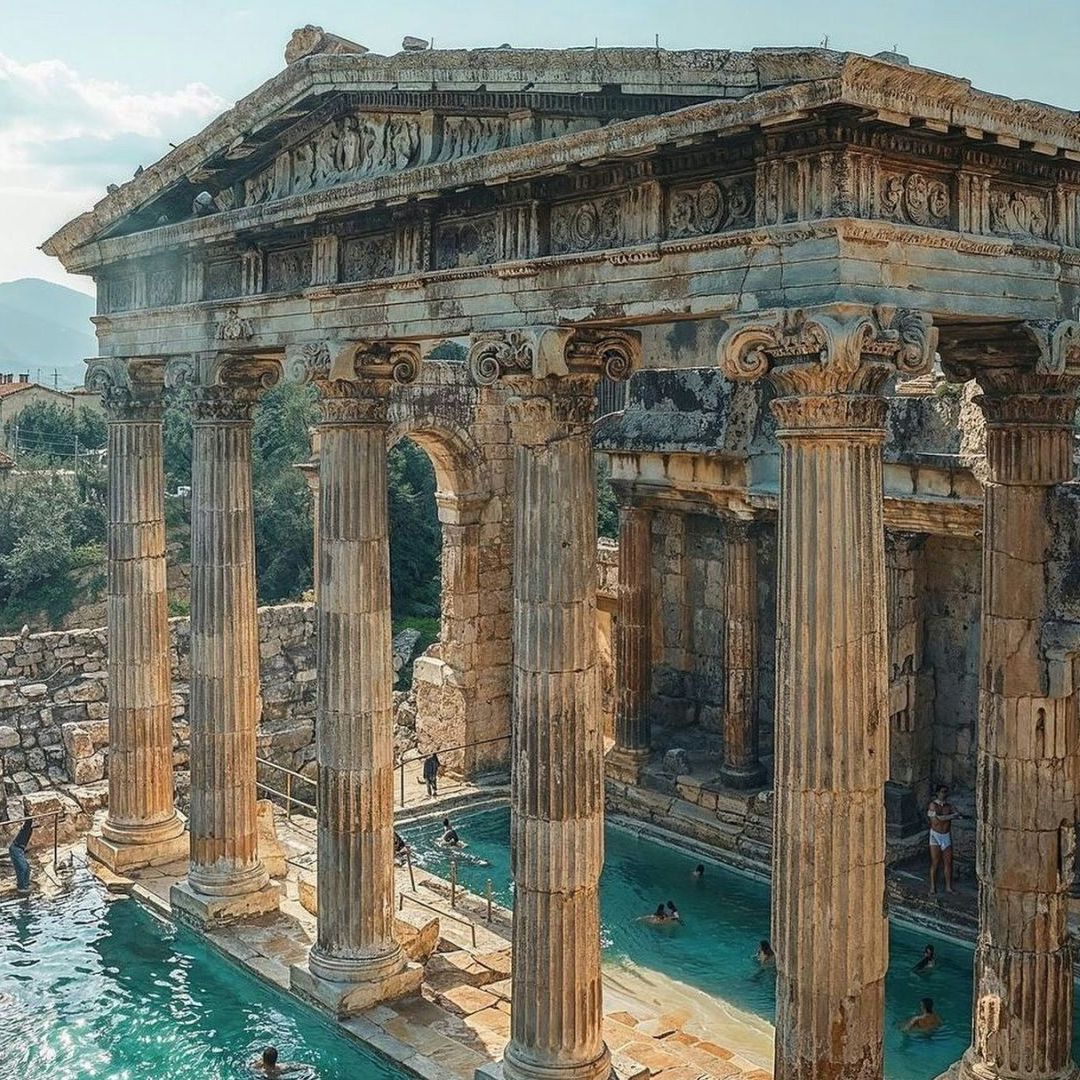}
        \caption{}
    \end{subfigure}
    \caption{Real (a-e) and generated (f-j) images from our introduced ITW-SM dataset.}
    \label{fig:itw-dataset}
\end{figure*}

AID methods can be categorized in five core categories: end-to-end supervised, vision language (VL) model-based, heuristic, reconstruction and zero-shot approaches.

End-to-end supervised models are trained on labeled datasets, allowing them to learn features that distinguish among real and generated images through supervised learning techniques. 
\citet{wang2020cnngeneratedimagessurprisinglyeasy} fine-tune a pre-trained ResNet50~\cite{he2015deepresiduallearningimage} on 20 different object classes of real images from LSUN~\cite{yu2016lsunconstructionlargescaleimage} and ProGAN~\cite{karras2018progressivegrowinggansimproved} images. To better capture generative artifacts, some methods focus on local image patches~\cite{chai2020makesfakeimagesdetectable} or combine local and global feature analysis~\cite{ju2022fusinggloballocalfeatures}. Another approach avoids downsampling in the early network layers to better preserve generative inconsistencies~\cite{gragnaniello2021gangeneratedimageseasy}. Building on this,~\citet{corvi2022detectionsyntheticimagesgenerated} fuse networks trained separately on GAN- and diffusion-based images to enhance generalization.

Vision-language approaches effectively distinguish images using features from models like CLIP~\cite{radford2021learning}. \citet{ojha2024universalfakeimagedetectors} adapt the protocol of~\cite{wang2020cnngeneratedimagessurprisinglyeasy} by using CLIP as a feature extractor rather than training a ResNet50. Other works improve text-to-image diffusion detection by integrating captions for joint analysis via CLIP's multimodal embeddings~\cite{sha2023defakedetectionattributionfake}, or by extracting representations from both intermediate and final encoder layers (RINE)~\cite{koutlis2024leveragingrepresentationsintermediateencoderblocks}.

Heuristic methods exploit predefined rules and known structural discrepancies between real and synthetic images. LGrad~\cite{tan2023learning} uses gradients from pre-trained CNNs as artifact representations, while \citet{tan2023rethinkingupsamplingoperationscnnbased} detect synthetic content by analyzing neighboring pixel dependencies introduced by upsampling in GANs and VAEs. More recently, AIDE~\cite{yan2025sanitycheckaigeneratedimage} classifies images by combining low-level texture statistics with high-level semantic embeddings.

Reconstruction approaches compare original and reconstructed image variants to highlight areas with artifacts that deviate from a learned distribution. To this end, DIRE~\cite{wang2023dirediffusiongeneratedimagedetection} measures the discrepancy between an input image and its reconstructed version generated by a pre-trained ablated diffusion model~\cite{dhariwal2021diffusionmodelsbeatgans}, utilizing these differences to train a ResNet50 classifier. In contrast, AEROBLADE~\cite{ricker2024aerobladetrainingfreedetectionlatent} avoids classifier training and directly uses reconstruction errors from a latent diffusion model's autoencoder, noting that generated images are reconstructed more accurately. Recently, SPAI~\cite{karageorgiou2024anyresolutionaigeneratedimagedetection} introduced a spectral detection method to learn the spectral distribution of real images in a self-supervised manner and identify generation artifacts via reconstruction similarity.

Zero-shot approaches detect generated images without being explicitly trained on such content. Early methods in this category model real image distributions via embedding perturbations~\cite{he2024rigidtrainingfreemodelagnosticframework} or pixel-wise reconstruction errors~\cite{cozzolino2024zeroshotdetectionaigeneratedimages}. More recently, vision-language models (VLMs) formulate detection as a multimodal reasoning or visual question answering task. Advanced techniques, including forensically-guided instructions and chain-of-thought reasoning, significantly boost zero-shot accuracy~\cite{kachwala2025taskaligned, promptengineered2025}, while soft prompt-tuning~\cite{chang2025antifakeprompt, keita2024fidavl} adapts VLMs for unified detection and source attribution. Despite these advances, recent evaluations reveal that VLMs' zero-shot performance consistently degrades over time against rapidly evolving generative models~\cite{chrysidis2026synthetic}. Furthermore, adapting such massive networks to continuous online distribution shifts incurs prohibitive computational costs. \looseness=-1

While the above works have significantly advanced the field of AID, a growing body of research has also investigated the robustness of AID methods to various perturbations~\cite{corvi2022detectionsyntheticimagesgenerated,corvi2023intriguingpropertiessyntheticimages,schinas2024sidbenchpythonframeworkreliably} and generative conditions beyond text~\cite{mareen2024tgif, giakoumoglou2025sagi, mareen2026tgif2}. These studies often reveal a significant drop in detection accuracy when models trained on unprocessed generated data are evaluated on generated images altered throughout their online lifecycle~\cite{karageorgiou2024evolution}. This suggests that learned features often overfit to specific generation artifacts and struggle against real-world distortions. To address this, our study sheds light on how common detector design choices interact with the diverse degradations encountered on social media platforms.

Recognizing the domain gap between lab-generated and in-the-wild content, recent datasets source images directly from online platforms. For example, Chameleon~\cite{yan2025sanitycheckaigeneratedimage} curates highly realistic, artist-refined images, but its focus on high-quality art fails to capture the degradation and noise typical of social media posts. TWIGMA~\cite{chen2023twigmadatasetaigeneratedimages} provides a large-scale collection of Twitter-scraped AI images, yet it is restricted to a single platform and lacks a balanced collection of authentic images. Broader benchmarks like AIGIBench~\cite{zeng2025aigibench} include social media samples to test robustness against distribution shifts but often lack semantic analysis and any provenance information. To facilitate realistic evaluation, we introduce a balanced, curated dataset sourced directly from verified pages across major social media platforms, including both authentic and AI-generated images. \looseness=-1
\section{Methodology}

\begin{figure*}
    \includegraphics[width=0.93\textwidth]{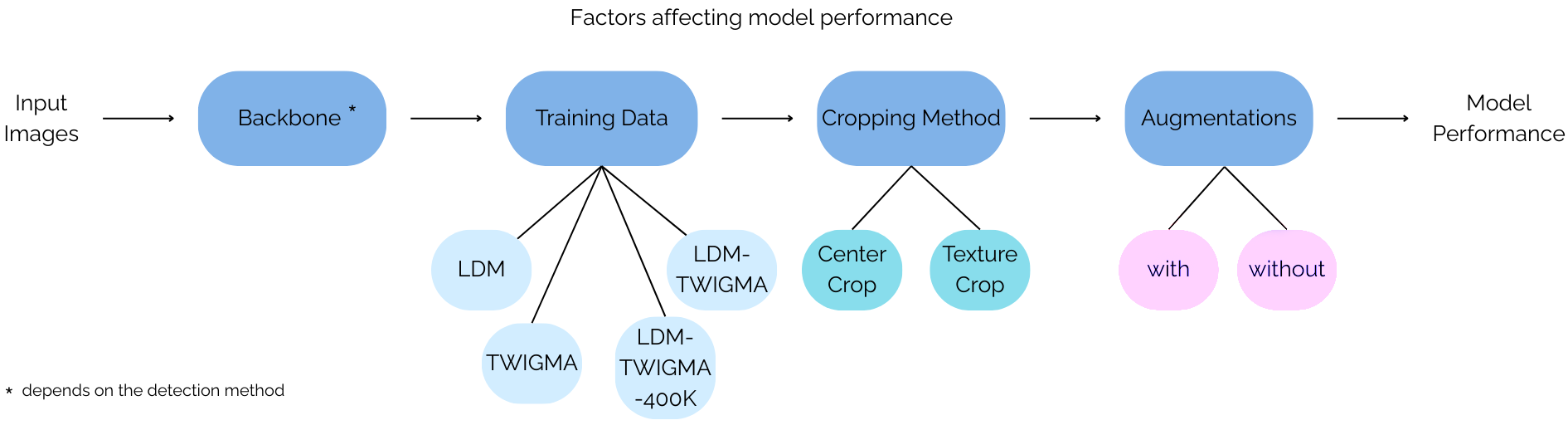}
    \caption{Framework for studying the factors impacting expected performance and generalization in AID models.}
    \label{fig:methodology}
\end{figure*}

\begin{table*}
\centering
\caption{Comparison between the proposed ITW-SM dataset and existing in the wild datasets.}
\renewcommand{\arraystretch}{0.9}
\setlength{\tabcolsep}{9pt}
\scalebox{0.95}{
\begin{tabular}{lccc}
\toprule
    \textbf{Key Differentiating Characteristics} & \textbf{ITW-SM} & \textbf{Chameleon} & \textbf{TWIGMA} \\
    \midrule
    AI-generated Images Source & Social media users & AI-painting communities & Twitter users \\
    Real Images Source & Verified social media accounts & Unsplash (photographers) & - \\
    Number of Social Media Platforms & 4 & 0  & 1 \\
    Resolution Range (All Images) & 0.1 - 45 Megapixels & 0.1 - 31 Megapixels & < 0.01 - 47 Megapixels  \\
    Size & 10,000 & 26,033 & 800,000 \\
    Intended Evaluation Focus & General in-the-wild robustness & Generalization to realistic AI & Analysis of AI art trends \\
\bottomrule
\end{tabular}
}
\label{tab:dataset_comparison}
\end{table*}

\begin{figure}
    \centering
    \includesvg[width=\columnwidth]{images/comparison_topics_grouped.svg}
    
    \caption{Topic distribution in web-collected datasets.}
    \label{fig:topics_comparison}
\end{figure}

\begin{figure}
    \centering
    \includesvg[width=\columnwidth]{images/comparison_resolution_boxplot.svg}
    
    \caption{Resolution distribution in web-collected datasets.}
    \label{fig:resolution_comparison}
\end{figure}

We propose an experimental framework (\cref{fig:methodology}) to systematically evaluate four key components: a) training data composition, b) backbone architecture, c) pre-processing cropping strategy, and d) training data augmentations.

\subsection{Training Data Composition}

Dataset composition significantly impacts AID robustness. While many studies rely on images from controlled environments or limited generators, such datasets fail to reflect real-world media complexity. Diverse training sets -- across both generative models and semantics -- enhance generalization to unseen architectures~\cite{wang2020cnngeneratedimagessurprisinglyeasy,corvi2022detectionsyntheticimagesgenerated}, though these benefits eventually level off~\cite{karageorgiou2024evolution}. Additionally, maintaining a balanced distribution of real and synthetic images is crucial for preventing biases that could hinder generalization. By training on datasets encompassing both benchmark and in-the-wild data, we assess how diversity influences generalization.

\subsection{Backbone Architectures for AID}

The backbone of a model dictates its ability to extract expressive features and detect subtle synthetic artifacts. Traditional convolutional neural networks (CNNs), such as ResNet~\cite{he2015deepresiduallearningimage} and EfficientNet~\cite{tan2020efficientnetrethinkingmodelscaling}, are widely used~\cite{cozzolino2021universalganimagedetection,gragnaniello2021gangeneratedimageseasy,corvi2022detectionsyntheticimagesgenerated,ju2022fusinggloballocalfeatures,mandelli2022detectinggangeneratedimagesorthogonal,Dogoulis_2023} for their strong spatial feature extraction. Recently, foundational models like CLIP~\cite{radford2021learning} have been adopted~\cite{amoroso2024parentschildrendistinguishingmultimodal,ojha2024universalfakeimagedetectors,cozzolino2024raisingbaraigeneratedimage,koutlis2024leveragingrepresentationsintermediateencoderblocks} for their superior capacity to capture both low-level details and high-level semantics. We evaluate how effectively these architectures identify artifacts and generalize in the wild, including images of varied depicted topics, lighting, resolution, and post-processing.

\subsection{Cropping Strategies}

Cropping directs models to particular image regions without relying on resizing, which risks erasing subtle high-frequency generation traces via interpolation~\cite{corvi2022detectionsyntheticimagesgenerated}. While center and random cropping remain the most common approaches, alternatives like 10-cropping (evaluating the center, four corners, and horizontal flips) have been utilized during inference to enhance performance~\cite{koutlis2024leveragingrepresentationsintermediateencoderblocks}. Additionally, texture-based cropping~\cite{konstantinidou2025texturecropenhancingsyntheticimage} targets high-frequency regions (e.g., edges, fine textures), demonstrating superiority over resizing and previous cropping methods. We systematically evaluate each strategy under a consistent pipeline to isolate its impact on detection performance.

\subsection{Data Augmentations}

Data augmentations simulate real-world distortions to improve model generalization. Common AID augmentations include compression (e.g., JPEG, WebP) to mimic online sharing artifacts, geometric transformations (cropping, rotating, flipping) for framing robustness, and noise/filtering techniques (Gaussian noise, blurring, sharpening) to build resilience against post-processing. Building on prior findings linking augmentation to detector robustness~\cite{wang2020cnngeneratedimagessurprisinglyeasy, mandelli2022detectinggangeneratedimagesorthogonal}, we investigate whether these augmentations enhance robustness when evaluated on heavily processed images.

\section{ITW-SM Dataset}

To meet the needs of our evaluation, we introduce the \textbf{In The Wild - Social Media Dataset (ITW-SM)}, specifically designed to reflect the complexity and diversity of online media content.

\subsection{Dataset Composition}

The ITW-SM dataset comprises \textbf{10,000 images}, evenly split between real and AI-generated ones. Real images are collected from verified, trusted accounts across four popular platforms: Facebook, Instagram, LinkedIn, and X. These were chosen to represent a diverse range of online content and image characteristics. Highlighting the real-world diversity, the images cover a wide range of topics and resolutions. Synthetic images are sourced from public accounts known to consistently share AI-generated content, including artists and communities that openly post images created with such tools. A comparison with previously introduced datasets that were also sampled from the web is presented in \cref{tab:dataset_comparison}.

\begin{figure*}[tb]
\includegraphics[width=0.93\textwidth]{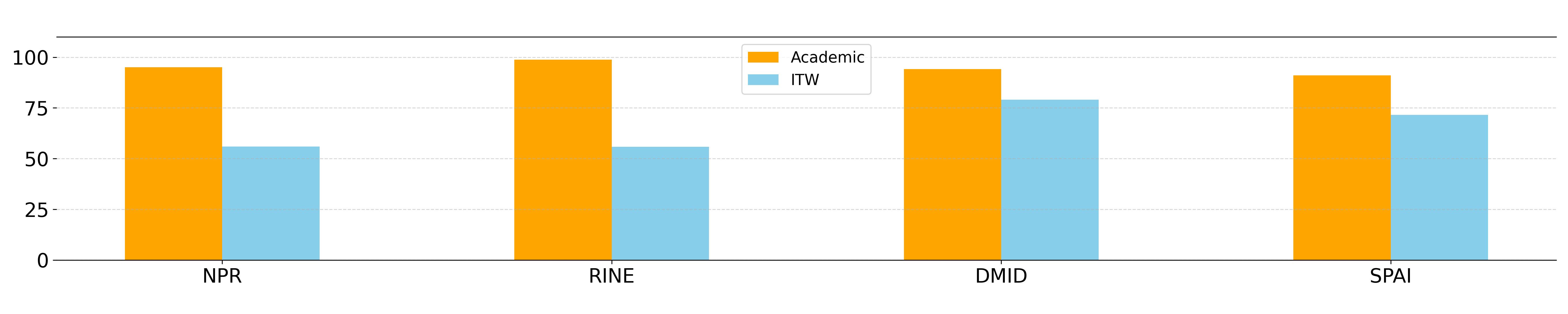}
    \vspace{-1.5em}
    \caption{Detection performance (AUC) on benchmark data (\textit{Academic}), as reported in original papers, and in the wild (\textit{ITW}).}
    \label{fig:academic}
\end{figure*}

\begin{table*}
  \caption{Performance (AUC/AP) of RINE with different backbones. Remaining components fixed to ``LDM+TWIGMA (1.2M)'', ``Texture cropping'', ``With augmentations''. For each training configuration, we explore a hyperparameter grid to achieve optimal performance, where $\xi$ is the contrastive loss factor, $q$ denotes the index of the network’s layer, $d$ is the output dimensionality of the backbone and $d'$ the output dimensionality of the projected feature space. Best values are highlighted in bold.}
  \label{tab:rine-backbones}
  \scalebox{0.95}{
  \setlength{\tabcolsep}{7.5pt}
\renewcommand{\arraystretch}{0.9}
  \begin{tabular}{lc|cccc|ccc|c}
    \toprule
    \multirow{2}{*}{\textbf{Model}} & \multirow{2}{*}{\textbf{Training Data}} & \multicolumn{4}{c|}{\textbf{Hyperparameters}} & \multicolumn{4}{c}{\textbf{Detection Performance (AUC / AP)}} \\
    \cmidrule(lr){3-6} \cmidrule(lr){7-10} 
          & & $\xi$ & $q$ & $d$ & $d'$ & \textbf{Synthbuster} & \textbf{Chameleon} & \textbf{ITW-SM} & \textbf{Average} \\
    \midrule
    CLIP L/14~\cite{radford2021learning} & 400M & 0.2 & 2 & 1024 & 1024 & 96.98 / 97.33 & 82.25 / 81.34 & 96.53 / 96.98 & 91.92 / 91.88 \\
    OpenCLIP L/14~\cite{Cherti_2023} & 2B & 0.2 & 2 & 1024 & 128 & 74.82 / 81.11 & 85.86 / 83.03 & 90.01 / 91.54 & 83.56 / 85.23 \\
    CLIP H/14~\cite{Cherti_2023} & 2B & 0.1 & 4 & 1280 & 256 & 97.02 / 81.71 & 81.22 / 76.98 & 90.56 / 91.81 & 89.60 / 83.50 \\
    BLIP2~\cite{li2023blip2bootstrappinglanguageimagepretraining} & 129M & 0.2 & 1 & 1408 & 1408 & \textbf{99.37} / \textbf{99.48} & 86.58 / \textbf{86.28} & 96.49 / 96.97 & 94.15 / 94.24 \\
    DINO-V2-L/14~\cite{oquab2024dinov2learningrobustvisual} & 142M & 0.8 & 1 & 1024 & 512 & 99.14 / 99.18 &  \textbf{87.33} / 85.51 & \textbf{98.23} / \textbf{98.50} & \textbf{94.90} / \textbf{94.40} \\
    \bottomrule
\end{tabular}
  }
\end{table*}

\begin{table}
\centering
\caption{Training data configurations.}
\renewcommand{\arraystretch}{0.83}
\scalebox{0.95}{
\begin{tabular}{lcccc}
\toprule
    \multirow{2}{*}{\textbf{Dataset}} & \multicolumn{2}{c}{\textbf{LDM Subset}} & \multicolumn{2}{c}{\textbf{TWIGMA Subset}} \\
    \cmidrule(lr){2-3} \cmidrule(lr){4-5}
    & \textbf{Real} & \textbf{Gen.} & \textbf{Real} & \textbf{Gen.} \\
    \midrule
    LDM & 200K & 200K & - & - \\
    TWIGMA  & - & - & 600K & 600K \\
    LDM+TWIGMA (400K) & 200K & 200K & 200K & 200K \\
    LDM+TWIGMA (1.2M)  & 200K & 200K & 600K & 600K \\
\bottomrule
\end{tabular}
}
\label{tab:datasets}
\end{table}

\begin{table*}
  \caption{Performance (AUC/AP) of detection methods when trained on different data configurations. Rest components fixed to ``DINO-V2-L/14'', ``Texture cropping'', ``With augmentations''. Best values per approach are highlighted in bold.}
  \label{tab:training-data}
  \setlength{\tabcolsep}{15pt}
\renewcommand{\arraystretch}{0.87}
  \scalebox{0.95}{
  \begin{tabular}{lcccc}
    \toprule
    & \textbf{Synthbuster} & \textbf{Chameleon} & \textbf{ITW-SM} & \textcolor{gray}{\textbf{Average}} \\
    \midrule
    \midrule
    \textbf{NPR} & & & \\
    \midrule
    LDM  & 55.95 / 58.47 & 53.04 / 44.57 & 54.85 / 55.92 & \textcolor{gray}{54.61 / 52.99} \\
    TWIGMA & 69.79 / 71.67 & 60.54 / 53.96 & \textbf{80.00} / \textbf{81.63} & \textcolor{gray}{\textbf{70.11} / \textbf{69.09}} \\
    LDM+TWIGMA (400K) & 63.51 / 66.56 & \textbf{63.29} / \textbf{55.77} & 43.59 / 52.66 & \textcolor{gray}{56.80 / 58.33} \\
    LDM+TWIGMA (1.2M) & \textbf{71.31} / \textbf{74.86} & 56.29 / 49.89 & 77.54 / 77.73 & \textcolor{gray}{68.38 / 67.49} \\
    \midrule
    \midrule
    \textbf{Gemma 3 IT 27B} & & & & \\
    \midrule
     zero-shot & \textbf{70.61} / \textbf{70.54} & \textbf{85.01} / \textbf{79.56} & \textbf{84.04} / \textbf{75.50} & \textcolor{gray}{ \textbf{79.89} / \textbf{75.20} } \\
    \midrule
    \midrule
    \textbf{DMID} & & & & \\
    \midrule
    LDM  & 81.26 / 82.79 & 54.69 / 55.03 & 79.96 /81.87 & \textcolor{gray}{71.97 / 73.23}\\
    TWIGMA & 90.27 / 90.15 & 82.97 / 78.51 & 91.55 / 92.03 & \textcolor{gray}{88.26 / 86.90} \\
    LDM+TWIGMA (400K) & 89.23 / 89.2 & 77.81 / 72.85 & 88.04 / 85.95 & \textcolor{gray}{85.03 / 82.67} \\
    LDM+TWIGMA (1.2M) & \textbf{92.4}/ \textbf{91.65} & \textbf{83.71} / \textbf{79.33} & \textbf{92.26} / \textbf{92.58} & \textcolor{gray}{\textbf{89.46} / \textbf{87.85}} \\
    \midrule
    \midrule
    \textbf{RINE} & & & & \\
    \midrule
    LDM  & 95.04 / 95.39 & 63.94 / 52.79 & 83.52 / 83.07  & \textcolor{gray}{80.83 / 77.08} \\
    TWIGMA & 97.77 / 97.96 & \textbf{89.03} / \textbf{86.69} & 97.66 / 98.07 & \textcolor{gray}{94.82 / 94.24} \\
    LDM+TWIGMA (400K) & \textbf{99.61} / \textbf{99.66} & 86.51 / 85.12 & 97.90 / 98.20 & \textcolor{gray}{94.67 / 94.33}  \\
    LDM+TWIGMA (1.2M) & 99.14 / 99.18  & 87.33 / 85.51  & \textbf{98.23} / \textbf{98.50}  & \textcolor{gray}{\textbf{94.90} / \textbf{94.40}} \\
    \midrule
    \midrule
    \textbf{SPAI} & & & & \\
    \midrule
    LDM   & 93.87 / 94.17 & 72.28 / 61.28 & 78.93 / 80.17 & \textcolor{gray}{81.69 / 78.54} \\
    TWIGMA & - & - & - & \textcolor{gray}{ - } \\
    LDM+TWIGMA (400K) & \textbf{98.81} / \textbf{99.02} & 89.38 / 87.26 & 97.59 / 97.80 & \textcolor{gray}{ \textbf{95.26} / 94.69 } \\
    LDM+TWIGMA (1.2M)& 97.45 / 98.00 & \textbf{90.21} / \textbf{88.51} & \textbf{98.10} / \textbf{98.35} & \textcolor{gray}{ 95.25 / \textbf{94.95} } \\
    \bottomrule
  \end{tabular}
  }
\end{table*}

\subsection{Collection Procedure}

Data collection was performed using a custom crawler that respects the terms of service of each platform. For each platform real content was scraped from verified accounts to ensure authenticity. AI-generated content was scraped from user accounts and community pages dedicated to sharing AI-generated visuals. All images were saved in their original resolution, preserving native compression artifacts. A multi-stage filtering pipeline was employed to ensure quality and consistency. First, images with heavy text overlays, watermarks, or non-photographic content (e.g., memes, screenshots) were removed. Then duplicates were eliminated using similarity scores and all samples were manually reviewed to verify label correctness. Moreover, as we wanted to maintain the distribution of AI-generated content shared online, and not merely make a dataset difficult to human evaluators, we avoided discarding images whose semantics may help to reveal whether they are generated or not. Fig.~\ref{fig:itw-dataset} illustrates some indicative samples of ITW-SM.

To better understand the composition of ITW-SM, we compare its topic and resolution distributions against existing web-sampled datasets, like Chameleon. For topic analysis, we define a taxonomy of 14 broad semantic categories (e.g., People, Nature, Art, Text/Memes) and perform zero-shot classification using OpenCLIP ViT-B/32~\cite{Cherti_2023}, assigning images based on maximum text-image cosine similarity. As \cref{fig:topics_comparison} illustrates, topic distributions differ significantly between social media platforms and online painting communities. Additionally, ITW-SM exhibits a substantially broader resolution range, particularly for real images (\cref{fig:resolution_comparison}).
\section{Experimental Setup}
\subsection{Datasets}

We train our detection models using both lab-controlled and in-the-wild data. For lab-controlled data, we use the LDM Training Dataset~\cite{corvi2022detectionsyntheticimagesgenerated}, a widely adopted benchmark that represents the architectural baseline for modern latent diffusion models~\cite{rombach2022high}. For in-the-wild data, we utilize TWIGMA~\cite{chen2023twigmadatasetaigeneratedimages}, a large-scale collection of web-sourced AI-generated images, combined with an equal number of real images from OpenImages~\cite{kuznetsova2020open}. This combination provides a robust foundation for evaluating generalization without over-fitting to the absolute newest generative architectures.

To bridge the gap between controlled experiments and real-world scenarios, we evaluate the models on three distinct datasets: Synthbuster~\cite{bammey2023synthbuster}, a highly-controlled benchmark containing 9k images generated by 9 models and 1k uncompressed real images from RAISE~\cite{ductien2015raise}; Chameleon~\cite{yan2025sanitycheckaigeneratedimage}, an in-the-wild dataset comprising over 11k high-fidelity AI and 14k real images from online creative communities; and our novel ITW-SM dataset, curated from four social media platforms to capture a diverse, real-world image distribution.\looseness=-1

\subsection{AID Models}

For our experiments, we utilize one representative model from each of the main categories in \cref{sec:related} (end-to-end supervised, VL model-based, heuristic, reconstruction and zero-shot).

\begin{itemize}
    \item \textbf{DMID}~\cite{corvi2022detectionsyntheticimagesgenerated} fuses the logits of two ResNet50 models independently trained on GAN- and diffusion-generated images. Both models utilize intense data augmentation and avoid down-sampling in their first layers.
    \item \textbf{RINE}~\cite{koutlis2024leveragingrepresentationsintermediateencoderblocks} leverages representations from intermediate CLIP Transformer blocks~\cite{radford2021learning} to capture both fine-grained details and high-level semantics. A trainable weighting module determines block importance, and a lightweight network maps features into a forgery-aware vector space.
    \item \textbf{NPR}~\cite{tan2023rethinkingupsamplingoperationscnnbased} detects artifacts introduced by generative up-scaling layers. It utilizes neighboring pixel relationships to train a detector focusing on the local pixel interdependencies caused by these up-sampling operators.
    \item \textbf{SPAI}~\cite{karageorgiou2024anyresolutionaigeneratedimagedetection} learns the real image distribution via masked spectral learning and frequency reconstruction. It detects out-of-distribution generated images using spectral reconstruction similarity, and employs spectral context attention to capture subtle inconsistencies across varying resolutions.
    \item \textbf{Gemma 3 IT 27B}~\cite{gemmateam2025gemma3technicalreport} is a zero-shot method that treats detection as a visual question-answering task as in~\cite{chrysidis2026synthetic}. Employed prompts constrain the output to strictly ``AI'' or ``REAL''.
    To ensure deterministic, single-token responses, we apply a 0.1 temperature and a 32-token maximum output.
\end{itemize}

\subsection{Evaluation Protocol}

To ensure a fair comparison of different AID methods, we retrain each model following the training details provided in their respective papers. In total, our experiments required more than 1000 GPU hours. While we adhere to the original training strategies of each detection approach, we make controlled adjustments to the backbones (when applicable), datasets, preprocessing methods, and augmentations to align with our experimental setup. To ensure computational feasibility across all methods, we apply similar resource constraints during training.
\section{Results}

\begin{table*}
  \caption{Performance (AUC/AP) of detection methods when trained using different cropping methods. Rest components fixed to ``DINO-V2-L/14'', ``LDM+TWIGMA (1.2M)'', ``With augmentations''. Best values per approach are highlighted in bold.}
  \label{tab:cropping}
\renewcommand{\arraystretch}{0.85}
  \setlength{\tabcolsep}{17pt}
  \scalebox{0.95}{
  \begin{tabular}{lcccc}
    \toprule
    & \textbf{Synthbuster} & \textbf{Chameleon} & \textbf{ITW-SM} & \textcolor{gray}{\textbf{Average}} \\
    \midrule
    \textbf{DMID} & & & & \textcolor{gray}{} \\
    \midrule
    Center cropping & 89.41 / 89.58 & 63.2 / 58.98 & 83.74 / 80.06 & \textcolor{gray}{78.78 / 76.21}  \\
    Texture cropping & \textbf{92.4}/ \textbf{91.65} & \textbf{83.71} / \textbf{79.33} & \textbf{92.26} / \textbf{95.58} & \textcolor{gray}{\textbf{89.46} / \textbf{87.85}}  \\
    \midrule
    \textbf{RINE} & & & & \\
    \midrule
    Center cropping & 94.65 / 94.65 & 84.08 / 82.7 & 95.04 / 95.75 & \textcolor{gray}{91.26 / 91.03} \\
    Texture cropping & \textbf{99.14} / \textbf{99.18} & \textbf{87.33} / \textbf{85.51} & \textbf{98.23} / \textbf{98.50} & \textcolor{gray}{\textbf{94.90} / \textbf{94.40}}  \\
    \midrule
    \textbf{NPR} & & & & \\
    \midrule
    Center cropping & \textbf{69.79} / \textbf{71.67} & 60.54 / 53.96 & \textbf{80.00} / \textbf{81.63} & \textcolor{gray}{\textbf{70.11} / \textbf{69.09}} \\
    Texture cropping & 64.08 / 65.78 & \textbf{62.34} / \textbf{54.24} & 76.03 / 76.98 & \textcolor{gray}{67.48 / 65.67} \\
    \bottomrule
  \end{tabular}%
  }
\end{table*}

\begin{table*}
  \caption{Performance (AUC/AP) of detection methods, trained on ``LDM+TWIGMA (1.2M)'' dataset, using different augmentations. Rest components fixed to ``DINO-V2-L/14'', ``Texture cropping''. Best values per approach are highlighted in bold.}
  \label{tab:augmentations}
  \setlength{\tabcolsep}{13pt}
  \begin{tabular}{lcccc}
    \toprule
    & \textbf{Synthbuster} & \textbf{Chameleon} & \textbf{ITW-SM} & \textcolor{gray}{\textbf{Average}} \\
    \midrule
    \textbf{DMID} & & & & \\
    \midrule
    Without augmentations & 76.53 / 66.47 & 76.21 / 69.36 & 82.20 / 81.42 & \textcolor{gray}{78.31 / 72.42} \\
    With augmentations & \textbf{92.4}/ \textbf{91.65} & \textbf{83.71} / \textbf{79.33} & \textbf{92.26} /\textbf{92.58} & \textcolor{gray}{\textbf{89.46} / \textbf{87.85}} \\
    
    \midrule
    \textbf{RINE} & & & & \\
    \midrule
    Without augmentations & 93.63 / 95.02 & \textbf{92.16} / \textbf{90.24} & 93.70 / 94.03 & \textcolor{gray}{93.16 / 93.10}\\
    With augmentations & \textbf{99.14} / \textbf{99.18} & 87.33 / 85.51  & \textbf{98.23} / \textbf{98.50} & \textcolor{gray}{\textbf{94.90} / \textbf{94.40}} \\
    
    \midrule
    \textbf{NPR} & & & & \\
    \midrule
    Without augmentations & \textbf{72.35} / \textbf{69.61} & 60.80 / 48.49 & 68.92 / 64.09 & \textcolor{gray}{67.36 / 60.73}\\
    With augmentations & 64.08 / 65.78 & \textbf{62.34} / \textbf{54.24} & \textbf{76.03} / \textbf{76.98} & \textcolor{gray}{\textbf{67.48} / \textbf{65.67}} \\
    
    \midrule
    \textbf{SPAI} & & & & \\
    \midrule
    Without augmentations & 94.80 / 94.83 & 82.73 / 81.02 & 91.46 / 91.75 & \textcolor{gray}{89.66 / 89.20} \\
    With augmentations & \textbf{97.45} / \textbf{98.00} & \textbf{90.21} / \textbf{88.51} & \textbf{98.10} / \textbf{98.35} & \textcolor{gray}{ \textbf{95.25} / \textbf{94.95} } \\
    \bottomrule
  \end{tabular}%
\end{table*}

Our findings confirm a key limitation previously discussed: while most methods achieve strong results on curated benchmark datasets, their performance degrades significantly when applied to in-the-wild AI-generated images, as illustrated in \cref{fig:academic}. 
To better understand the impact of each component in isolation, we vary one factor at a time while keeping the others fixed. We then compare the results from the original implementations, as reported in the respective papers, with those obtained through our re-evaluation.

\subsection{Backbone}

Recognizing the significance of selecting effective backbones for AID tasks, we conducted an ablation study centered on the CLIP-based RINE model trained on LDM data. We selected RINE because it exemplifies a representative, simple and competitive approach in the recent literature of AID methods. In our study, we replaced the CLIP L/14 backbone~\cite{radford2021learning} in RINE with various alternative vision encoders. The backbone is characterized by three main components: the architecture itself, the pretraining objective, and the dataset used during pretraining. 
In this ablation, we primarily focus on Vision Transformer (ViT)-based architectures, so the key variations lie in the pretraining objectives and the diversity of the pretraining datasets. The encoders tested include:

\begin{itemize}
    \item 
    \textbf{OpenCLIP L/14}~\cite{Cherti_2023}
    This open-source implementation of CLIP is designed to provide greater flexibility and transparency in large-scale VL modeling. OpenCLIP L/14 is trained on the LAION-2B dataset~\cite{schuhmann2022laion5bopenlargescaledataset} and also extends the CLIP architecture with improvements in training.
    \item \textbf{BLIP2}~\cite{li2023blip2bootstrappinglanguageimagepretraining}: This integrates frozen pre-trained image encoders with large language models (LLMs) by employing a lightweight 12-layer Transformer encoder in between, trained on a 129M image dataset introduced in~\cite{li2021alignfusevisionlanguage}, achieving state-of-the-art results on various VL tasks.
    \item 
    \textbf{CLIP H/14}~\cite{Cherti_2023}: This variant of the CLIP model employs advanced scaling techniques to enhance performance across different applications. It is a larger and more powerful version of CLIP L/14 trained also on  LAION-2B~\cite{schuhmann2022laion5bopenlargescaledataset}.
    \item 
    \textbf{DINO-V2-L/14}~\cite{oquab2024dinov2learningrobustvisual}: This is pretrained on large curated datasets without supervision. It incorporates an optimized training recipe, increased model scale, and a larger  curated dataset, LVD-142M~\cite{oquab2024dinov2learningrobustvisual}, along with a distillation process that enables smaller models to benefit from the capabilities of the most powerful ViT architecture.
\end{itemize}

We present the respective performance in \cref{tab:rine-backbones}. DINO-V2's superior performance likely stems from its self-supervised training focused purely on visual understanding, its ability to capture both low-level and semantic features robustly, and the scale and curation of its training data. Results further highlight the importance of quality pre-training data. One additional explanation is that CLIP-based methods' reliance on image-text alignment may introduce semantic shortcuts, emphasizing contextual relevance over fine-grained visual details. This can lead to representations that are less sensitive to low-level inconsistencies—such as texture aberrations or local artifacts—that are crucial for detecting AI-generated content and should be jointly considered with image semantics.

\begin{figure*}[htb]
\includegraphics[width=0.88\textwidth]{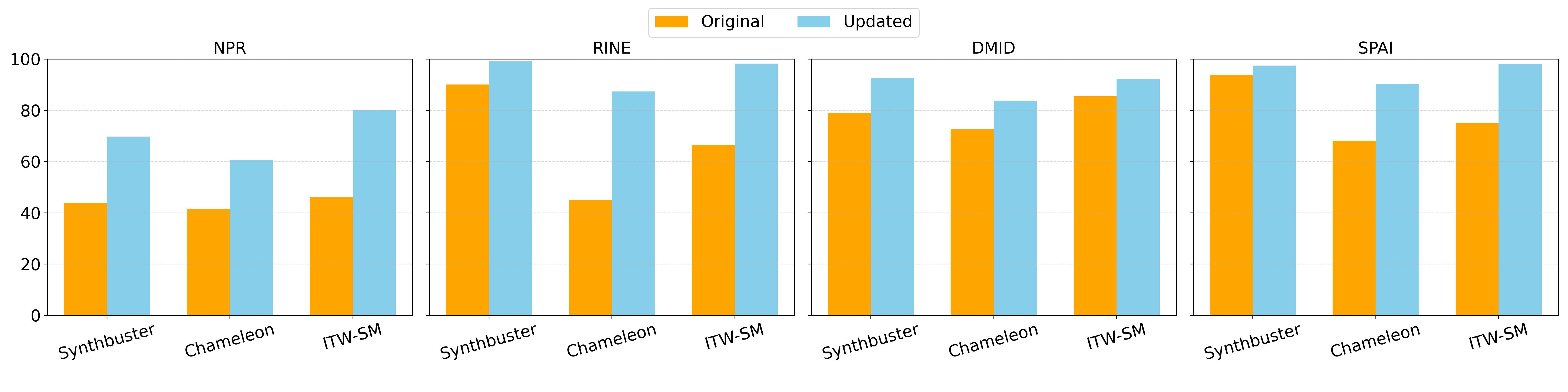}
    \vspace{-1.5em}
    \caption{Original and updated model performance (AUC).}
    \label{fig:comparison}
\end{figure*}

\subsection{Training Data}

We use the LDM training dataset~\cite{corvi2022detectionsyntheticimagesgenerated} and TWIGMA~\cite{chen2023twigmadatasetaigeneratedimages} to retrain our models. The former consists of 200K latent diffusion-generated images and 200K real images sourced from two public datasets: MS COCO~\cite{lin2015microsoftcococommonobjects} and LSUN~\cite{yu2016lsunconstructionlargescaleimage}. All images in this dataset are of low resolution.
We consider 600K AI-generated images from TWIGMA for training, while we use an equal number of real images  from the OpenImages dataset~\cite{kuznetsova2020open}. This includes both low- and high-resolution images. We combine these datasets to create four training datasets, as can be seen in \cref{tab:datasets}.

Based on our analysis in \cref{tab:training-data}, we first observe that the performance of all detection approaches benefits from training on data collected in the wild, even when considering lab-generated benchmarks, like  Synthbuster. However, approaches that heavily rely on pre-trained spaces, like RINE and SPAI, are only marginally affected by increasing the training data scale. Instead, the performance of the end-to-end supervised detector DMID, due to optimizing all its representations from scratch, benefits significantly more from such a scaling. Also, targeting specific generation artifacts, like in the case of NPR, significantly prevents an approach to benefit from more diverse training data.

\subsection{Cropping Method}

Inspired by the promising results of~\cite{konstantinidou2025texturecropenhancingsyntheticimage}, we adopt TextureCrop during training, randomly selecting one of the 10 crops per image to reduce overhead. It is important to note that the SPAI model is not included in this analysis, as it natively operates on patches of the original image rather than a single crop.

Based on our results, TextureCrop appears to significantly boost performance compared to center cropping on the DMID and RINE methods, as it enables them to process more informative image  and capture more robust generation traces, by targetting regions with high texture information. However, advanced cropping approaches can also compromise the ability of detectors that make strong assumptions about the generative artifacts. This is exemplified by NPR, of which the performance degrades when altering the expected cropping format.

\subsection{Augmentations}

By artificially expanding the diversity of the training set, augmentation techniques help the model recognize generative artifacts across varying conditions and scenarios. Our experimental results demonstrate that incorporating comprehensive data augmentation strategies improves on average the performance metrics (AUC/AP) of all methods across all three datasets, as seen in \cref{tab:augmentations}. Interestingly, for RINE, augmentations slightly degrade performance on Chameleon, suggesting that the applied augmentations may have introduced transformations that deviate from the types of distortions found in the dataset. For most methods, we used the augmentations mentioned in the corresponding papers. However, for the NPR model, since no intense augmentations were originally applied, we considered the augmentation pipeline of \cite{koutlis2024leveragingrepresentationsintermediateencoderblocks}.

\subsection{Comparison of Original and Updated Implementations}

To assess the impact of our proposed modifications, we compare the performance of the original implementations and the updated models after applying our changes. The obtained results  are presented in \cref{fig:comparison}, where we display the AUC values for each method before and after applying our updates. Our modifications yield significant improvements in AID performance, which attests to their efficacy, especially for tackling AID in the wild.
\section{Conclusion}

Our study highlights the critical challenges that AID models face in real-world applications and identifies key factors that influence detection performance. By analyzing the role of backbone architectures, training data composition, cropping methods, and data augmentations, we provide actionable insights for improving AID robustness. The findings demonstrate the need for more robust detection techniques that account for real-world variations and remain effective in practical settings. Such analysis should be conducted on any new model to be deployed in the wild, as different models exhibit different behaviors and may require tailored strategies for optimal performance.

\begin{acks}
We thank Zacharias Chrysidis for his invaluable assistance on late-stage experimentation with VL models. This work was funded by the Horizon Europe projects vera.ai (GA No. 101070093), AI-CODE (GA No. 101135437), and ELIAS (GA No. 101120237). Computational resources were provided by the National Infrastructures for Research and Technology GRNET and funded by the EU Recovery and Resiliency Facility.
\end{acks}


\bibliographystyle{ACM-Reference-Format}
\bibliography{main}

\end{document}